\documentclass[journal]{IEEEtran}
\usepackage{color}
\usepackage{times}
\usepackage{epsfig}
\usepackage{graphicx}
\usepackage{amsmath}
\usepackage{amssymb}
\usepackage{multirow}
\usepackage{floatrow}
\usepackage{mathrsfs}
\usepackage{array}
\usepackage{ifpdf}
\usepackage{algorithmic}
\usepackage{algorithm}
\usepackage{cite}
\usepackage{subcaption}
\usepackage{changes}
\makeatletter
\let\Changes@Markup@Deleted\@gobble
\makeatother
\usepackage{url}
\usepackage[colorlinks,
            linkcolor=red,
            anchorcolor=blue,
            citecolor=green 
            ]{hyperref}

\begin{document}
\title{Real-World Image Super-Resolution by Exclusionary Dual-Learning}

\author{ Hao Li\textsuperscript{\dag} , Jinghui Qin\textsuperscript{\dag}, Zhijing Yang, Pengxu Wei, Jinshan Pan, Liang Lin, Yukai Shi

\thanks{
 {\dag} The first two authors share equal contribution.
 }
 \thanks{
 Code is available at: \url{https://github.com/House-Leo/RWSR-EDL}
 }
}

\maketitle

\begin{abstract}
Real-world image super-resolution is a practical image restoration problem that aims to obtain high-quality images from in-the-wild input, has recently received considerable attention with regard to its tremendous application potentials. Although deep learning-based methods have achieved promising restoration quality on real-world image super-resolution datasets, they ignore the relationship between L1- and perceptual- minimization and roughly adopt auxiliary large-scale datasets for pre-training. In this paper, we discuss the image types within a corrupted image and the property of perceptual- and Euclidean- based evaluation protocols. Then we propose a method, Real-World image Super-Resolution by Exclusionary Dual-Learning (RWSR-EDL) to address the feature diversity in perceptual- and L1- based cooperative learning. Moreover, a noise-guidance data collection strategy is developed to address the training time consumption in multiple datasets optimization. When an auxiliary dataset is incorporated, RWSR-EDL achieves promising results and repulses any training time increment by adopting the noise-guidance data collection strategy. Extensive experiments show that RWSR-EDL achieves competitive performance over state-of-the-art methods on four in-the-wild image super-resolution datasets.

\end{abstract}


%
\IEEEpeerreviewmaketitle

\section{Introduction}
\IEEEPARstart{R}{eal-world} image super-resolution (RealSR) aims at restoring in-the-wild images collected from poor-quality sensors with unknown degraded kernels. Since RealSR realizes image restoration under real-world scenarios, it plays a remarkable role in many human-centric applications, such as mobile photo enhancement, automatic pilot, etc. 

Traditional single image super-resolution (SISR)~\cite{yan2021fine,jiang2020dual,tian2020coarse,li2020learning,shi2017structure,qin2019difficulty} obtains high-resolution images from low-quality ones with known and fixed degradation models (e.g., Gaussian blur followed by Bicubic downsampling). As the in-the-wild corrupted images own complicated degradation kernels, traditional SISR model exhibit limited capacity when applying to in-the-wild applications. To address this issue, RealSR constructs the real-world image pairs by incorporating poor- and high- quality sensors to take low- and high- quality images, respectively. Compared with manual downsample kernel, the degraded kernel in RealSR is inherently more complicated and closed to in-the-wild degradation. As the mobile platform has a limited optical sensor size and a large number of users, mobile photography enhancement is one of the most challenging applications for RealSR. Thus, RealSR prefers to entertain the human visual system. Typical evaluation protocols in traditional SR (e.g., PSNR and SSIM~\cite{ssim}) focus on pixel-level similarity and fail to reflect human perception well. Recently, various kinds of efforts~\cite{fid,lpips,lin2018hallucinated,chen2020knowledge,jiang2021degrade} are proposed to reflect the human visual system in image quality assessment. LPIPS~\cite{lpips} argues that the widely used pixel-to-pixel measuring methods (e.g., L2/PSNR, SSIM, and FSIM) are contrary to human perception when estimating the perceptual similarity of images. 

\begin{figure*}[t]
    \centering
 \vspace{-0.35cm}
 \begin{subfigure}{0.49\textwidth}
    \includegraphics[width=\textwidth]{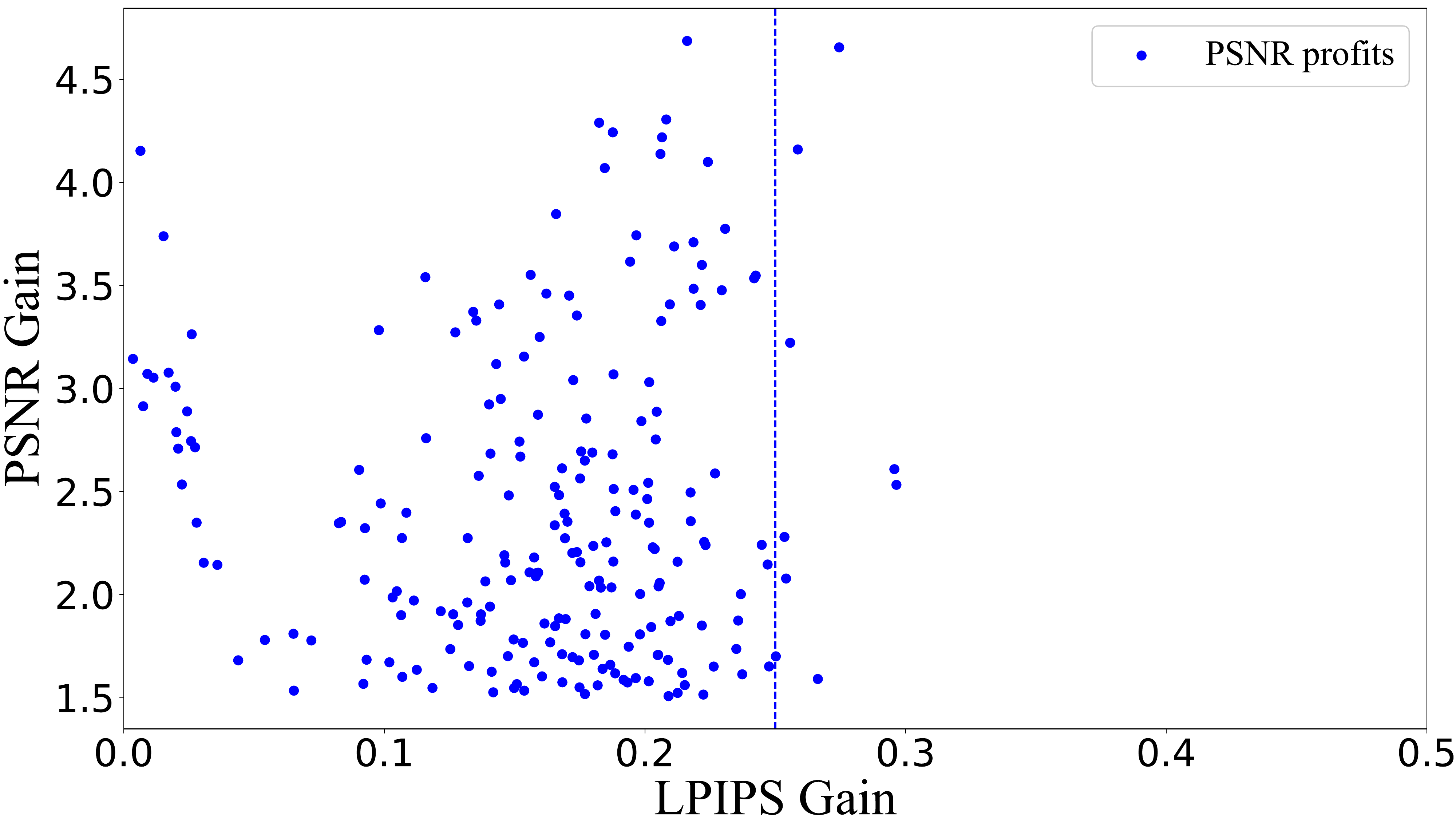}
    \caption{High PSNR Profits}
 \end{subfigure}
 \begin{subfigure}{0.49\textwidth}
    \includegraphics[width=\textwidth]{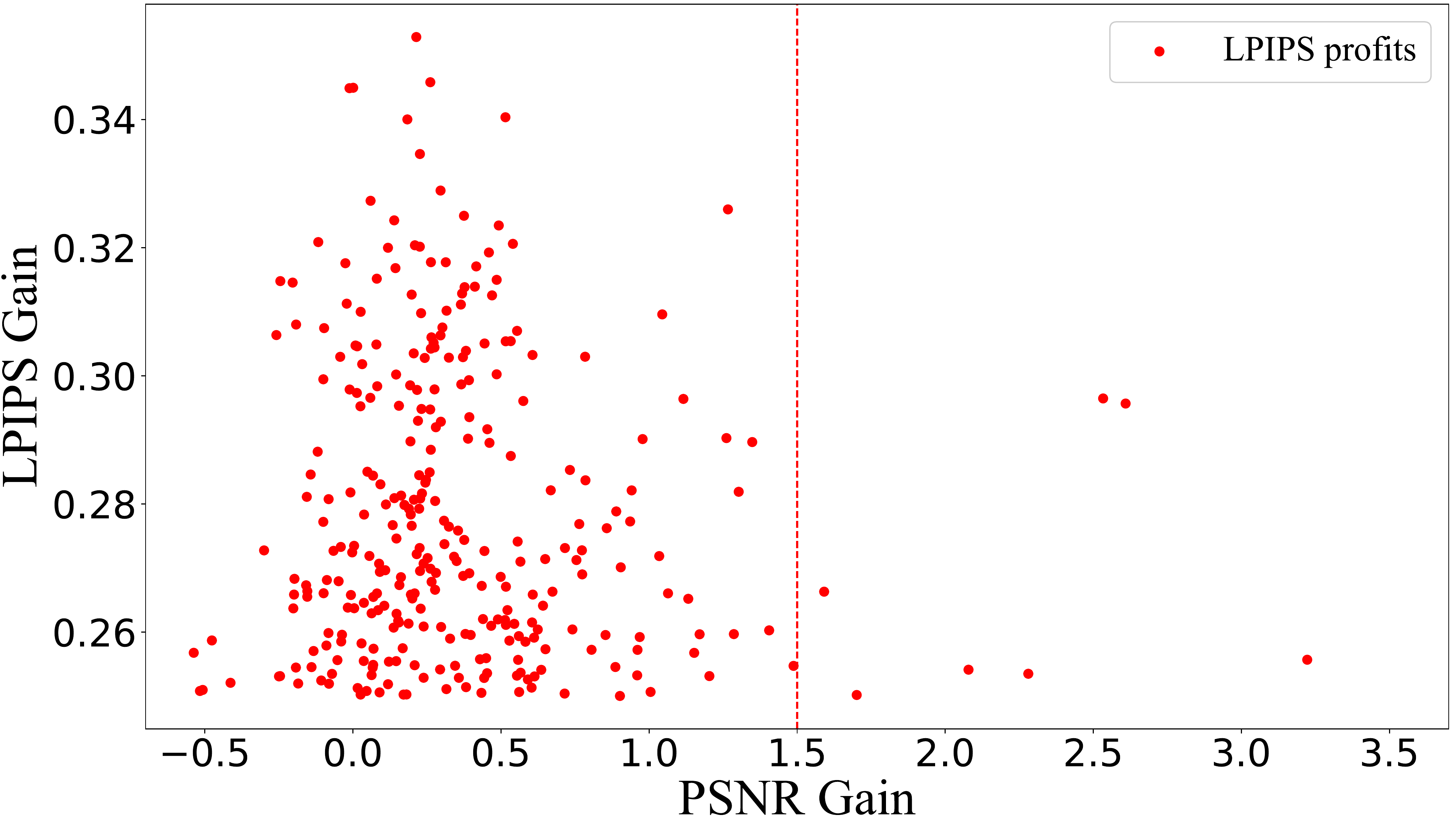}
    \caption{High LPIPS Profits}
 \end{subfigure}
 
  \begin{subfigure}{\linewidth}
    \includegraphics[width=\linewidth]{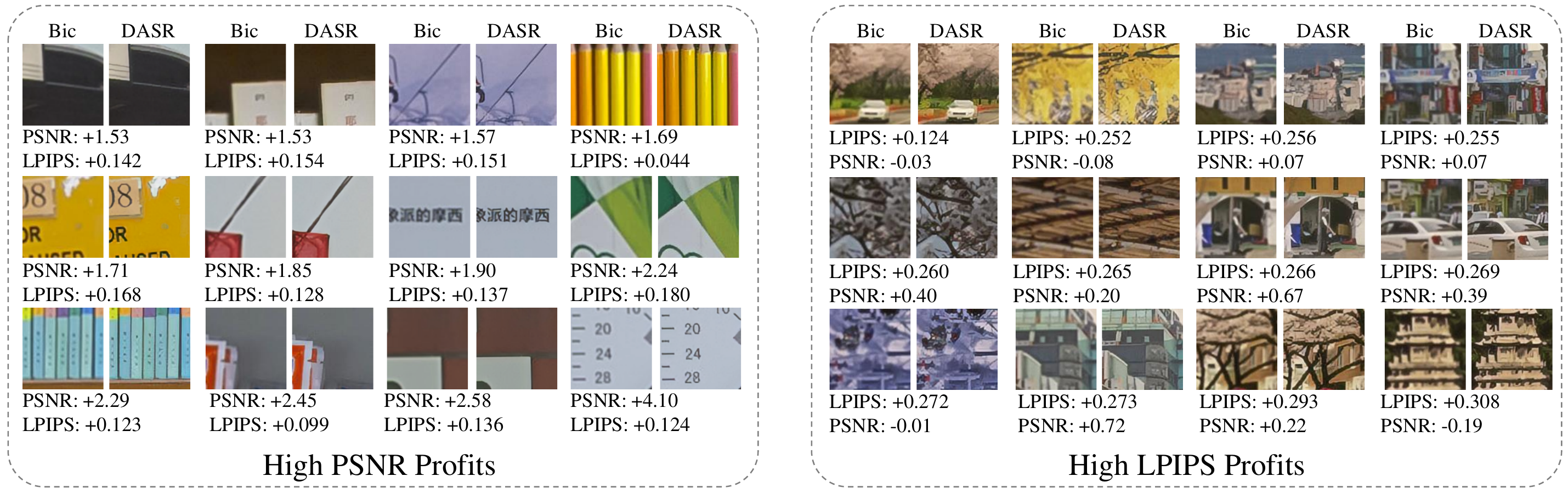}
    \caption{Visualization of high-profit results with PSNR and LPIPS index.}
 \end{subfigure}
    \caption{We obtain high-profit results of DASR~\cite{wei2020unsupervised} from the perspective of PSNR and LPIPS index. Its discernable that the left high PSNR gain examples($>> 1.5 dB$) obtain clear background, smooth structure and inconspicuous LPIPS promotion. However, the high LPIPS profits samples($>> 0.25$) contain complicated texture and a few artifacts, which also obtain less PSNR gain. Though PSNR and LPIPS are both used for evaluating image restoration, they fail to reach an agreement.}
    \label{fig:psnr_lpips}
\vspace{-3mm}
\end{figure*}

Recently, RealSR methods~\cite{AIM19,Lugmayr2020ntire,shi2020ddet} usually adopt LPIPS and PSNR as default evaluation metrics and achieve superior scores either on PSNR or LPIPS. As depicted in Fig.~\ref{fig:psnr_lpips}, high PSNR increment samples exhibit simple background, smooth structure, and relatively weak LPIPS increment. Besides, high LPIPS gain samples own complicated texture and less PSNR gain. On one hand, images with complex textures are easier to get high scores on LPIPS on account of psychophysical similarity measurements~\cite{lpips}, while the adversarial training is adept at generating artificial texture. On the other hand, Euclidean-based measurements inherently prefer L1 minimization. Although the metrics based on adversarial loss and L1-norm-based loss are able to generates the results with favorable both LPIPS and PSNR values, the prior methods still utilize the weighted ratio to simply joint them all.

Nevertheless, multiple datasets are another challenge for real-world image restoration frameworks. In RealSR task, many datasets~\cite{Lugmayr2020ntire,CameraSR,AIM19,realsr,chen2021cross,wei2020component} are proposed to address the real-world degradation and pixel displacement. Extensive training on multiple RealSR datasets is beneficial for generalization enhancement as well as performance improvement~\cite{wei2020component}. However, typical backbones~\cite{edsr,zhang2018image,zhang2020residual,jiang2020dual} require several days to be converged on one standard dataset~\cite{timofte2017ntire} with a single GPU. The incremental number of RealSR datasets would raise more requirements toward efficiency. Specifically, the training efficiency would suffer a heavier burden when auxiliary datasets are incorporated for pre-training. This motivates us to investigate a dataset distillation strategy for multiple dataset collaborative training. 

Previously, research communities pay considerable attention to model distillation as large-scale dataset plays a important role in training time consumption. For instance, DIV2K has 1,000 images with 2K resolution, NTIRE2020 and AIM2020 challenges provide 3450 images with 2K resolution, respectively. Though having large-scale training, especially when encountering data augmentation, can significantly improve performance, the following training time cost certainly receives an obvious increment.

To tackle the above challenge, we observe that a full image contains many image types, which lead to a large variance in structure signal. Inspired by this, we introduce a Dual-Learning strategy based on a exclusionary attention mechanism to facilitate different types of images to exhibit diverse representation under multi-task learning paradigms. To address training time consumption issue, we also explore noise extraction and blending mechanism across multiple datasets, and proposed an efficient data collection strategy. With the proposed data collection strategy, we certainly relax the training time growth problem while incorporating an auxiliary large-scale dataset for mix-training. Our contributions can be summarized as:

\begin{itemize}
\item We propose an efficient training paradigm(i.e., NGDC algorithm) that extra time consumption is relaxed when incorporating auxiliary datasets for training.
\item We proposed an exclusionary mask mechanism , namely RWSR-EDL, that brings significant improvement toward multi-loss training conditions in RealSR. 
\item We provide a comprehensive comparison on four challenging real-world SR benchmarks (e.g., AIM2019~\cite{AIM19}, {NTIRE2020}~\cite{Lugmayr2020ntire}, CameraSR~\cite{CameraSR}, RealSR~\cite{realsr}) to demonstrate that RWSR-EDL achieves a clear performance improvement on PSNR and LPIPS index both, and present a high-quality image restoration. 
\end{itemize}

\section{Related Work}
\textbf{Single Image Super Resolution.}
Deep learning-based SR methods~\cite{jiang2019atmfn,jiang2021rain} have achieved significant improvements over conventional SR methods on restoration quality. Among these methods, Dong et al. proposed the first CNN-based SR method called SRCNN~\cite{dong2014learning}, which utilizes a three-layer CNN to learn a nonlinear mapping between LR and HR image pairs. Lim et al. proposed EDSR~\cite{edsr}, using simplified residual blocks for training SR model, which can achieve great improvement over restoration quality. Luo et al.~\cite{luo2021ebsr} proposed a Feature Enhanced Pyramid Cascading and Deformable convolution (FEPCD) module to align multiple low-resolution burst images in the feature level. Zhang et al. proposed a residual in residual (RIR)~\cite{zhang2018image} to address the difficulty of training a deep SR network with a channel attention mechanism to improve the representation ability of the SR network. Wang et al. proposed ESRGAN~\cite{wang2018esrgan}, which introduced a Residual-in-Residual Dense Block (RRDB) into the generative network. Furthermore, FASRGAN~\cite{yan2021fine} explored a novel pixel-level discriminator for a fine-grained generative adversarial learning and obtains interesting results. Hu et al.~\cite{hu2020meta} address the simplicity of the downscaling kernel in image SR by adopting multiple degradation metrics, and achieves promising results on complex corrupted images. Recently, many dual-way neural network~\cite{jiang2020dual} are applied to image restoration, however, they only address the traditional image SR problem and neglect the complexity of real-world cases. Mei et al.~\cite{csnln} proposed the first Cross-Scale Non-Local (CS-NL) attention module and a powerful Self-Exemplar Mining (SEM) cell into deep neural networks. Jiang et al.~\cite{jiang2020hierarchical} proposed a hierarchical dense connection network (HDN) for SR, achieved the improvements of both reconstruction performance and efficiency.

Although the above deep learning-based SR methods bring significant improvement over SISR, they cannot be generalized well on real-world images as their assumed the known and fixed degradation process does not hold for real-world images.

\textbf{Real-World Super Resolution.}
Recently, Real-world super-resolution attracts considerable attention due to its distinguished practicality. Different from the traditional image SR that generally focuses on simple and uniform synthesized degradation, RealSR needs to handle complicated and rough degradation in real-world cases. To address the above challenges, Shocher et al. proposed zero-shot super-resolution (ZSSR)~\cite{zeroshot}, they realize an unsupervised CNN-based SR method by training an image-specific SR network with internal data rather than employing external data. Bell-Kligler et al. proposed KernelGAN~\cite{kernelgan}, to generate down-sampling kernel from label images by adopting kernel estimation GAN, which used in ZSSR for degraded kernel estimating. Fritsche et al. proposed the DSGAN model~\cite{fritsche2019frequency} to generate LR-HR pairs and then apply ESRGAN-FS on corresponding generated images. Pang et al. proposed FAN~\cite{fan} to extract the different frequency components of the LR images, which can be used to recover the HR images by preserving more high-frequency details. Wei et al. proposed a DASR framework~\cite{wei2020unsupervised} by calculating the domain distance between LR images and real images with the domain-gap aware training and domain-distance weighted supervision. Ji et al.~\cite{Ji_2020_CVPR_Workshops} proposed an RWSR model based on ESRGAN by using kernel estimation and noise injection.

However, the above methods only employ a single structure network to obtain the enhanced image, they ignore that the multi-branch network can learn more diverse image features with various prior knowledge. In contrast, we use a dual-learning refinement module and a exclusionary mask generator to further extract the diverse representation. With our exclusionary dual-learning strategy, we obtain the final enhanced image with rich high-frequency details, which indeed address the complicated degradation in real-world super-resolution.

Since the ground-truth label may absent in some image enhancement applications, some no-reference image quality assessment (IQA) metrics~\cite{liu2014no,mittal2012making,blau20182018} are proposed. Spatial–Spectral Entropy-based Quality (SSEQ)~\cite{liu2014no} is capable of assessing the quality of a distorted image across multiple distortion categories. Naturalness Image Quality Evaluator (NIQE)~\cite{mittal2012making} uses the multivariate Gaussian (MVG) model to fit the quality-aware features extracted from images. And Perception Index (PI)~\cite{blau20182018} combines the no-reference image quality measures of Ma et al.~\cite{ma2017learning} and NIQE~\cite{mittal2012making} to get the score.

\textbf{Noise Modeling Based Denoising.} 
Reducing the effect of noise is a critical issue in real-world image restoration. Recently, some noise modeling-based approaches have been proposed to address the real-world noise estimation and reduction problem. Lebrun et al. proposed Noise Clinic (NC)~\cite{Noiseclinic}, a method to estimate the noise model dependent on signal and frequency followed by using non-local Bayes (NLB)~\cite{NLB} . Zhang et al. proposed FFDNet~\cite{FFDNet}, a non-blind Gaussian denoising network that can obtain charming results on real-world noise cases, nevertheless, it requires manual intervention to select noise level.

However, real-world image noise is indeed distinct from Addictive White Gaussian Noise(AWGN) and the network trained by the handcraft degradation gives a poor performance when applied to complex real-world noise. Guo et al. proposed CBDNet~\cite{CBDNet}, which is composed of two sub-networks, that is noise estimation and non-blind denoising net. They use a noise modeling method that is able to generate realistic noise and also adopt real-world clean images for paired-wised training. Chen et al. proposed GCBD~\cite{GCBD} to extract the smaller image with clear background in noisy images, and then incorporating GAN to generate more fake noise samples for training denoising CNN network. Anwarf et al. proposed RIDNet~\cite{RIDNet}, a single-state denoising network with feature attention for real-world noise reduction.

Different from the above approaches, our RWSR-EDL is free from noise modeling, instead, we make use of inherent noise sampling strategy towards the real-world images to construct paired training data, which can eliminate the biases introduced by noise modeling. Also, we incorporate the NGDC strategy that effectively distills a large-scale auxiliary dataset to obtain target real-world noise type. With NGDC strategy, the performance is clearly improved without training time increment.

\begin{figure*}[t]
	\begin{center}
		\includegraphics[width=0.98\linewidth]{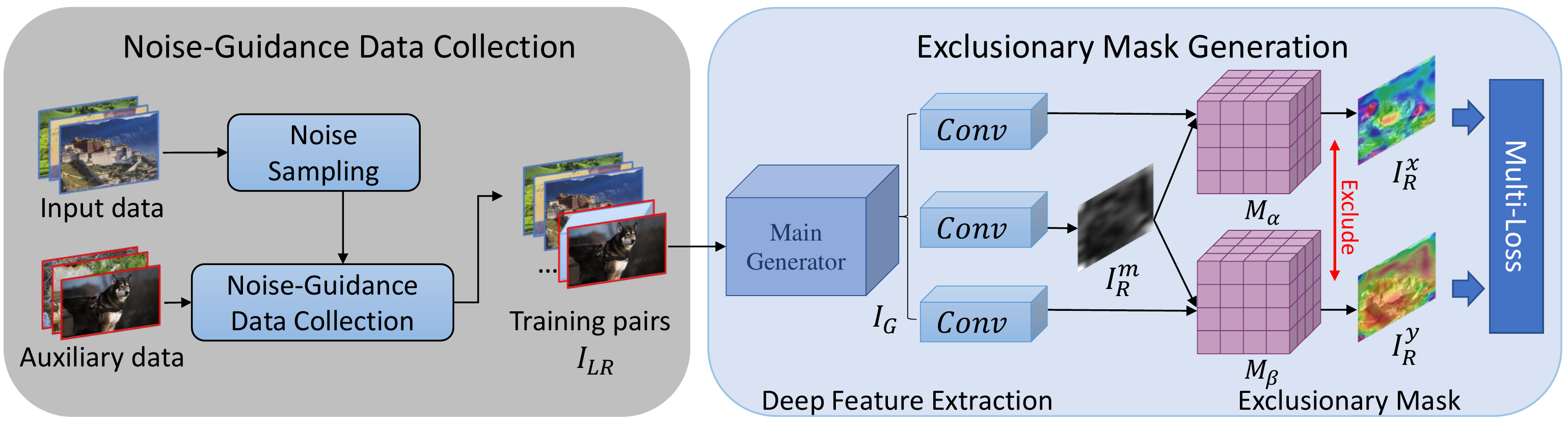}
	\end{center}
	\caption{Network architecture of RWSR-EDL, which consists of two components: 1) Noise-Guidance Data Collection for an efficient training in multiple large-scale RealSR datasets; 2) Exclusionary Mask Generator for relaxing multi-loss optimization in RealSR.}
	\label{fig:network}
	\vspace{-3mm}
\end{figure*}

\section{Methodology}
\subsection{Overview}
 Traditional single image super-resolution(SISR) aims at restoring a high-quality image $I_{HR}$ from a low-quality image $I_{LR}$.
In traditional SISR, the $I_{LR}$ is synthesized from $I_{HR}$ with a down-scaling operation:
\begin{equation}
I_{LR} = (I_{HR} \otimes K_{Gauss}) \Downarrow_{\Bbbk} .
\end{equation}
where $K_{Gauss}$ is a Gaussian blur kernel and $\Downarrow_{\Bbbk}$ denotes the image degradation procedure with downscale factor $\Bbbk$. To obtain a low-quality $I_{LR}$, $\Downarrow_{\Bbbk}$ typically adopts a bicubic-based downsampling algorithm. In contrast to traditional SISR, real-world super-resolution aims to address real-world image degeneration metrics by capturing $I_{HR}$ and $I_{LR}$ with different quality optical sensors and resolution settings where the degraded 
metrics(e.g., $\Downarrow_{\Bbbk}$) and blur kernel are unknown. Therefore, $\left \{ I_{LR},I_{HR} \right \}$ pairs inherently has different resolution properties in the real-world data collection procedure. 

To improve the quality of real-world images and attack existed noises and artifacts in real-world images, we propose a novel single image super-resolution framework to take precedence over learned image features. As shown in Figure~\ref{fig:network}, there are two key components in our RWSR-EDL: 1) noise-guidance data collection for an efficient training in multiple large-scale RealSR datasets; 2) exclusionary mask generator for relaxing multi-loss optimization in RealSR. In our method, we first enforce the LR images and ground-truth images yield similar noise distribution by embedding random noise into LR images. The intermediate HR images generated by the main feature extractor will be entered into the exclusionary dual-network. Finally, our framework learns diverse intermediate representations adaptively to pursue high PSNR and LPIPS scores by incorporating exclusionary mask and multi-loss optimization. 

\subsection{Main Generator}
Since the goal of our work is to improve the perceptual quality of SR images under the real-world setting, we adopt an effective image enhancement backbone for feature extraction, which consists of 23 residual-in-residual dense blocks (RRDBs~\cite{wang2018esrgan}) and incorporate paired-wised training. Given $I_{LR}$, we generate intermediate HR image $I_G$ with the main generator as follows:
\begin{equation}
I_G = Generator(I_{LR}).
\end{equation} 

\vspace{-4mm}
\subsection{Exclusionary Mask Generation}
To obtain a better super-resolving image by further improving the intermediate HR image $I_G$, we employ a deep feature extraction, which consists of two branches with the same network design but various initialization metrics to fully explore diverse feature representation. The three branches both contain ResBlocks~\cite{residual_net} which has two 3$\times$3 convolutional layers. The first convolutional layer with 3 input channels and 64 output channels is followed by a ReLU activation function while the second convolutional layer has 64 input channels and 3 output channels. To this end, each ResBlock in feature extraction phase can be formulated as follows:
\begin{equation}
I_{R}^x = I_{G} + f_{RB}^{x}(I_{G}).
\label{euq:I_G}
\end{equation}
where $x$ represents the $x_{th}$ ResBlock in the three branch, $f_{RB}$ indicates the ResBlock, $I_{G}$ and $I_{R}^x$ means the input and output of ResBlock, respectively. Similar to Eqa.~\ref{euq:I_G}, we obtain $[I_{R}^x , I_{R}^y. I_{R}^m]$ w.r.t three parallel branches.

As the refined images $I_{R}^x$ and $I_{R}^y$ are obtained, we deploy a soft-mask generator to demonstrate the exclusionary dual-learning. More specifically, $I_{R}^x$ and $I_{R}^y$ are optimized with various loss functions to pursuing PSNR and LPIPS promotion both by applying exclusionary masks. Let $I_{R}^m$ represent the output of the second branch, we apply a softmax operator on $I_{R}^m$ to normalize all feature values into 0.0 to 1.0 as an adaptive mask. Specifically, a soft-mask for channel index can be generated as follows:

\begin{equation}
M_{\alpha} =\operatorname{softmax}\left(I_{R}^m\right)=\frac{\exp \left(z_{n}\right)}{{\sum_{n=1}^{N}} \exp \left(z_{n}\right)}
\label{eqa:alpha}
\end{equation}

where $n$ and $z_{n}$ are channel number and specific value of $I_{R}^m$ with channel $n$ index, respectively. To this end, we obtain the $M_{\alpha}$ for exclusionary dual-learning.

\subsection{Loss Function}
Since the exclusionary dual-learning is proposed relax the conflicts of perceptual- and L2- based optimizations, we first briefly introduce the the losses we used in this paper, including Pixel Loss, Perceptual Loss, and Adversarial Loss.

\textbf{Pixel Loss.} we use L1 loss, which is a widely used loss function for general image restoration, to train our generator to recover as much effective pixel as possible. The L1 loss is defined by the manhattan distance between the reconstructed image $I_{SR}$ and the ground-truth image $I_{HR}$ as follows:
\begin{equation}
\mathcal{L}_{pix}(I^{i}_{SR},I^{i}_{HR})=\frac{1}{N} \sum_{i=1}^{N}\left\|I^{i}_{HR}-I^{i}_{SR}\right\|_{1}
\end{equation}
where $N$ is the samples of training set.

\begin{figure*}[t]
    \centering
    \includegraphics[width=0.49\textwidth]{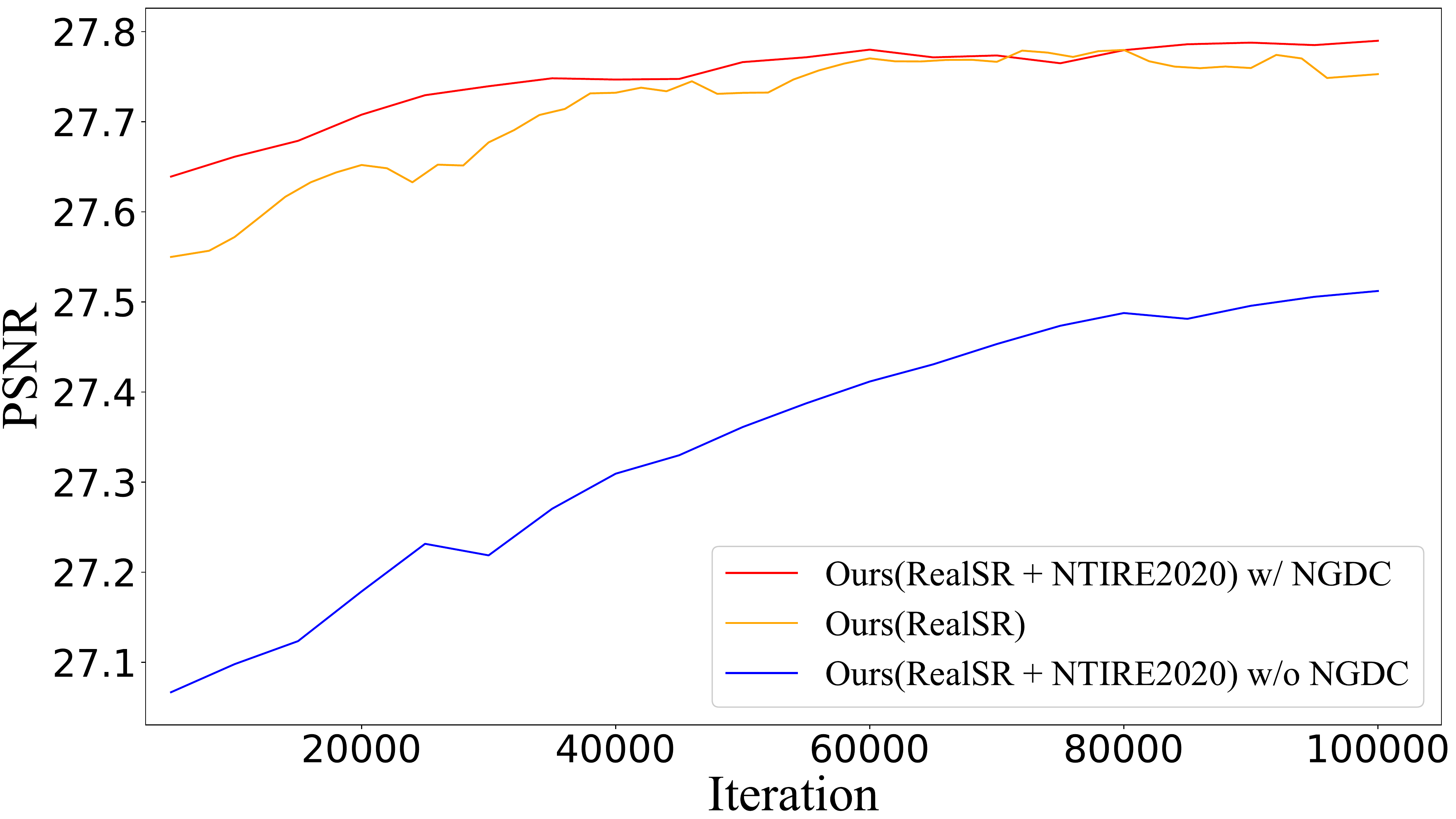} 
    \includegraphics[width=0.49\textwidth]{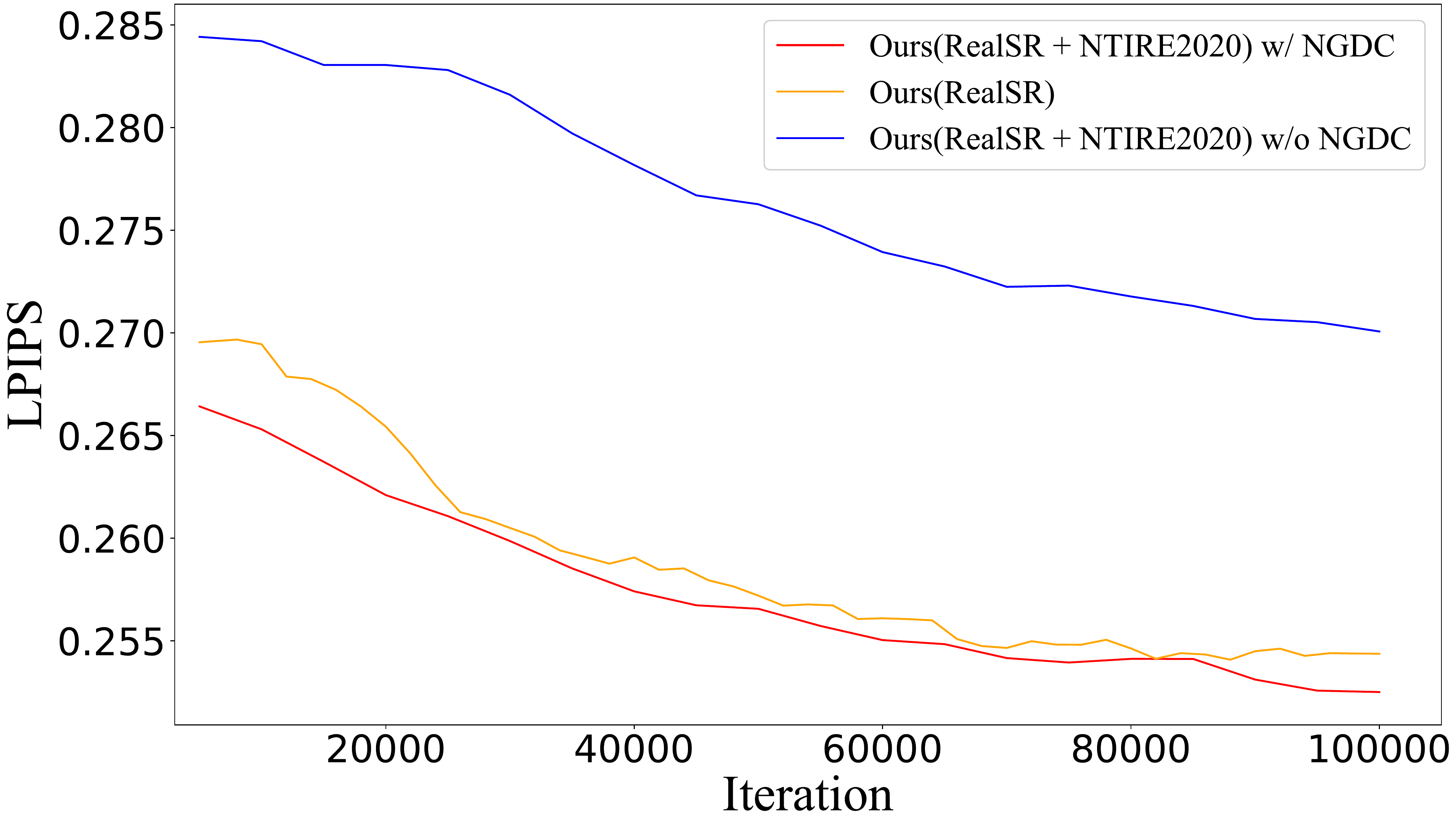}
    \caption{PSNR/LPIPS curves of different settings on RealSR~\cite{realsr} with training iteration index. With NGDC strategy, although we incorporate auxiliary large-scale dataset(i.e., NTIRE2020~\cite{Lugmayr2020ntire}) for training, \textbf{the training time receives no gain while the performance is clearly improved.} }
    \label{fig:curves}
\vspace{-3mm}
\end{figure*}

\textbf{Perceptual Loss.} To further enhance the high-frequency features (such as edges) in the SR image, we deploy a perceptual loss based on feature space. Specially,  we extract the features of $I_{SR}$ and $I_{GT}$ with a pre-trained VGG-19, and compute their loss as follows:
\begin{equation}
\mathcal{L}_{per}(I^{i}_{SR},I^{i}_{HR})=\frac{1}{N} \sum_{i=1}^{N}\left\|VGG_{19}(I^{i}_{HR})-VGG_{19}(I^{i}_{SR})\right\|_{1}
\end{equation}
where $N$ is the samples of training set and $VGG_{19}$ denotes a pre-trained VGG-19 model~\cite{vgg}.

\textbf{Adversarial Loss.} We also deploy adversarial loss to enhance the SR image's texture to make it more realistic. The adversarial loss is defined as follows:

\begin{align*}
\mathcal{L}_{adv}(&I^{i}_{SR},I^{i}_{HR}) = \frac{1}{N}\sum_{i=1}^{N}\{-E[\log (1-\sigma(D(I^{i}_{HR})-  \\
&E(D(I^{i}_{SR}))))] - E[\log (\sigma(D(I^{i}_{SR})-E(D(I^{i}_{HR}))))]\}.
\end{align*}
where $N$ is the samples of the training set, $\sigma$ represents a sigmoid function, and $D(\cdot)$ is the discriminator. In adversarial learning, patch discriminator takes advantage of typical used VGG-128 with some aspects. First, patch discriminator is a fully convolutional network, which is free from image size restriction by getting rid of the fully-connected layer. Second, local feature representation is excavated with the limited receptive field. We then apply patch discriminator instead of VGG-128 as $D(\cdot)$. 

\textbf{Exclusionary Dual-Learning.} As illustrated in Fig.~\ref{fig:psnr_lpips}, a region with high LPIPS profits often bears low PSNR increment. However, a typical utilization of $[\mathcal{L}_{adv}, \mathcal{L}_{per},\mathcal{L}_{pix}]$ in imageSR~\cite{srgan} is employ them together with weighted average:

\begin{equation}
\begin{aligned}
\widetilde{\mathcal{L}}_{all}=\alpha \mathcal{L}_{adv}(I_{SR},I_{HR})& + \beta \mathcal{L}_{per}(I_{SR} ,I_{HR}) \\
&+ \gamma \mathcal{L}_{pix}(I_{SR},I_{HR}),
\end{aligned}
\label{eqa:oriloss}
\end{equation}

where $[\acute{\alpha}, \acute{\beta}, \acute{\gamma}]$ are empirical weight factors. $\widetilde{\mathcal{L}}_{all}$ incorporates all losses with fixed weight factors for optimization and ignores the diversity of image types in term of perceptual- and L1-norm-based metrics. Contrary to the Equ.~\ref{eqa:oriloss}, we adopt exclusionary masks in multiple losses training to avoid domain-conflicts and obtain more accurate feature representation:

\begin{equation}
\begin{aligned}
\mathcal{L}_{all}=\mathcal{L}_{adv}(M_{\alpha} \cdot I_R^x,I_{HR})& + \mathcal{L}_{per}(M_{\alpha} \cdot I_R^x + M_{\beta} \cdot I_R^y,I_{HR}) \\
&+ \mathcal{L}_{pix}(M_{\beta} \cdot I_R^y,I_{HR}),
\end{aligned}
\label{eqa:loss}
\end{equation}
where $\cdot$ is a matrix dot operation. Meanwhile, we obtain $M_{\alpha}$ with Equ.~\ref{eqa:alpha}, and $M_{\beta}$ is obtained from $M_{\alpha} + M_{\beta} = 1 $ to promise the two masks have exclusionary property. 

In Eqa.~\ref{eqa:loss}, $M_{\alpha}$ enforces partial region of $I_R^x$ optimized with $\mathcal{L}_{adv}$ and avoid the distraction of $\mathcal{L}_{pix}$. Simultaneously, $M_{\beta}$ performs spatial attention on $I_R^y$ for $\mathcal{L}_{pix}$ optimization and free from $\mathcal{L}_{adv}$. The overall output $[I_R^x+I_R^y]$ is further smooth by $\mathcal{L}_{per}$ in Eqa.~\ref{eqa:loss}. With the proposed exclusionary dual-learning mechanism, we can demonstrate a fine-grained multi-loss optimization and achieve high PSNR and LPIPS profits both.

\emph{Difference to Spatial Attention Mechanism.} Compared with spatial attention mechanism~\cite{zhou2016learning,woo2018cbam}, which simply highlights the feature according to the high entropy information, the proposed exclusionary dual-learning mechanism enforces the branches capture diverse feature with exclusionary masks. More specific, the generated soft-mask utilizes a branch to obtain high-value feature representation as usual, then, another branch is enforced to capture extensive information. With this competitive mechanism, our model can learn complementary features at the same time and demonstrates promising results on complex real-world conditions. To this end, we address that different image types exhibit different signal properties with deeply learned representation and the adaptive mask is able to reconstruct better visual-quality SR images. In Ablation Study Section, we will show that using the adaptive mask is superior to using a plain dual-way neural network.

\subsection{Noise-Guidance Data Collection}
\textbf{Data Preparation.} 
As the real-world images inherently own a certain proportion of noise, which leads to the restored images contain spare artifacts, we apply a down-sampling to reduce the negative impact. Specifically, we incorporate a bicubic kernel $K_{bic}$ on the source image $I_{src}$ to implement noise remove and obtain the $I_{HR}$ :

\begin{equation}
\label{eqa:down1}
I_{HR} = (I_{src} \otimes  K_{bic})\Downarrow_{\Bbbk}.
\end{equation}
Similarly, we can obtain the corrupted image $I_{LR}$ with:
\begin{equation}
I_{LR} = (I_{HR} \otimes  K_{bic})\Downarrow_{\Bbbk}.
\end{equation}

\textbf{Noise Sampling.} 
The high-frequency signal among $I_{src}$ has a certain amount of drawback as the Eqa.~\ref{eqa:down1} reduces the pattern information. The noise distribution on $I_{LR}$, therefore, meets a significant change. In order to keep the $I_{LR}$ and $I_{src}$ has similar noise distribution, we directly sampling the noise in $I_{src}$ with a grid strategy. Specifically, a sliding window with a size of $s \times s$ and the stride is $s$ as well is used to capture smaller images. The mean and variance value of the captured patch is computed for sample sifting. 

Suppose variance and mean value of a fetched patches $n_i$ is $\sigma_i$ and $m_i$. We obtain the CharAcTeristic Interval (CATI) (i.e., $[ \sigma_{\theta_1},\sigma_{\theta_2}]$ and $[m_{\theta_1},m_{\theta_2}]$) by acquiring bottom $2\%$ patches from the perspective of variance. When the $\sigma_i$ and $m_i$ pertain to CATI, we regard $n_i$ as a noise patch, and joint it into the noise patch bank $N$. Otherwise, we regard the rest patch as noiseless patch.

\begin{figure}[t]
	\begin{center}
		\includegraphics[width=0.89\linewidth]{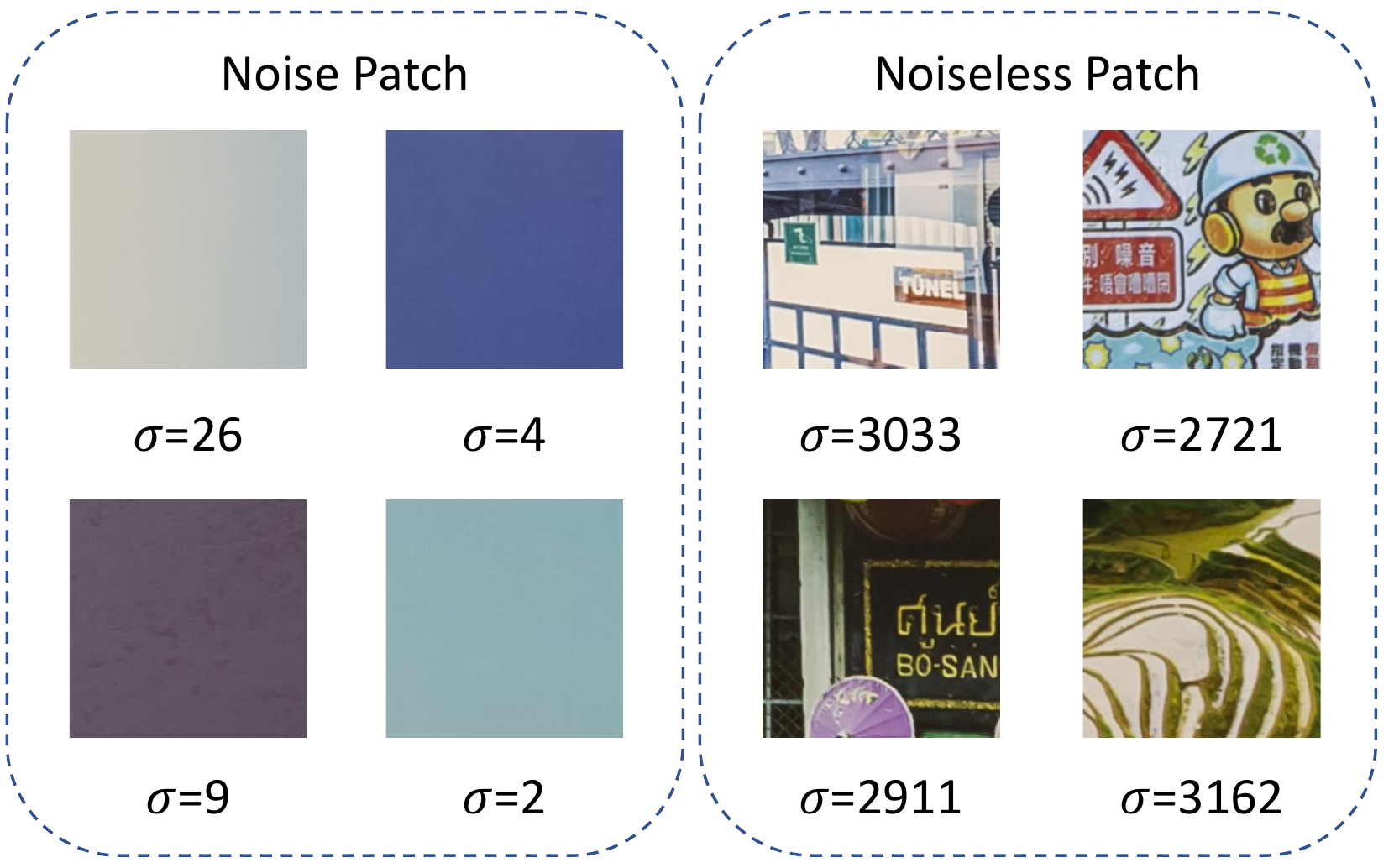}
	\end{center}
	\caption{Noise and Noiseless patch.}
	\label{fig:noise_comparison}
\end{figure}

As shown in Fig.~\ref{fig:noise_comparison}, the noiseless patch, which contains richer texture and detail, has larger $\sigma$. We regard these variances exceeded patch as noiseless patch along with two aspects: 1) Complex structure and texture may cover up the noise. 2) Noise signal collection across multiple datasets benefit from the clean background. To this end, the patch with smaller $\sigma$ and positive $m$, which enjoys plain content and help us purely extract the inherent noise, is collected into noise bank.

Then, we randomly fetch $n_i$ from the constructed $N$ and infuse $n_i$ into $I_{LR}$ to generate training input $I'_{LR}$ as:

\begin{equation}
\label{eqa:noise_infuse}
I'_{LR} = I_{LR} + n_i.
\end{equation}
As illustrated in Eq.~\ref{eqa:noise_infuse}, the training pairs become more complex and the noise resistance capability of our model is enhanced with the discriminative learning.

\textbf{Noise-Guidance Data Collection.} 
In inspired by the noise sampling phrase, we extend it to cross-dataset noise collection. Assuming the noise samples, which owns noise type similar to the target domain, is existed in an additional dataset. We aim at distilling auxiliary datasets and indexing similar noise domain image. More specifically, we collect a noisy bank $N_a$ from the target dataset $D_a$ by incorporating its CATI. Then, we apply CATI, which was obtained from $D_a$ already, to the auxiliary dataset $D_b$ as well. Notice here that $D_b$ is usually larger than $D_a$. For instance, suppose a source image $I_{src}^b$ in $D_b$ pertains to the CATI, we generate the corresponding $\left \{ I_{HR}^{b},I_{LR}^{b},N^{b} \right \}$ and merge them into $\left \{ I_{HR}^{a},I_{LR}^a,N^a \right \}$. Finally, the distilled training pairs $\left \{ I_{LR}^{a+b}, I_{HR}^{a+b},  N^{a+b} \right \}$. We sketch the overall algorithm in Alg.~\ref{alg1}. With NGDC, we can significantly reduce the data volume of auxiliary dataset as well as achieve better performance. 

{\emph{Difference to Impressionism.} As depicted in~\cite{Ji_2020_CVPR_Workshops}, Impressionism collects noise patches by the following rule: $\sigma(p)<v$, where $\sigma(\cdot)$ denotes the function to calculate variance in image patch $p$, and $v$ is value of a fixed threshold. However, $v$ is an empirical value and needs a manual search when using a different training set. To remedy this,  as demonstrated in Algorithm.~\ref{alg1}, we obtain the patches from the bottom 2\% variance by calculating the variance of all patches, and this rule can be applied to any training set. As shown in Fig.~\ref{fig:curves}, NGDC realizes an efficient training paradigm that achieves a faster convergence by collecting valuable noise patches across multiple training sets.}

\begin{algorithm}[t]
        \caption{Algorithm of NGDC}
        \begin{algorithmic}[1] 
        \label{alg1}
        \REQUIRE Target dataset $D_a$, auxiliary dataset $D_b$, noise bank $N_a=\varnothing$ and $N_b=\varnothing$
        
        \FOR{$I_i^a$ in $D_a$}
            \STATE  Compute $\sigma_i$ and $m_i$ by calculating $I_i^a$ 
            \STATE $I_{HR}^a = (I_{i}^a \otimes  K_{bic})\Downarrow_{\Bbbk} $
            \STATE $I_{LR}^a = (I_{HR}^a \otimes  K_{bic})\Downarrow_{\Bbbk} $
        \ENDFOR
        \STATE Construct $N_a$ from bottom $2\%$  $\sigma$ value images and compute corresponding CATI $\in$ $[\sigma,m]$
        \FOR{$I_i^a$ in $D_a$}
            \FOR {patch $p_j^a$ in $I_i^a$}
                \STATE Compute $\sigma$ and $m$ for $p_j^a$
                \IF { $ \left \{ (\sigma, m) \vert p_j^a \right \}  \in CATI$ }
                    \STATE ${N}_a = {N}_a + p_i^a$
                \ENDIF
            \ENDFOR
        \ENDFOR
        
        \STATE Construct $\left \{ I_{LR}^{a}, I_{HR}^{a},  N^{a} \right \}$
        \FOR{$I_i^b$ in $D_b$}
            \FOR {patch $p_j^b$ in $I_i^b$}
                \STATE Compute $\sigma$ and $m$ for $p_j^b$
                \IF { $ \left \{ (\sigma, m) \vert p_j^b \right \}  \in CATI$ }
                    \STATE ${N}_b = {N}_b + p_i^b$
                    \STATE $I_{HR}^b = (I_{i}^b \otimes  K_{bic})\Downarrow_{\Bbbk} $
                    \STATE $I_{LR}^b = (I_{HR}^b \otimes  K_{bic})\Downarrow_{\Bbbk} $
                \ENDIF
            \ENDFOR
        \ENDFOR
        \STATE Obtain $\left \{ I_{LR}^{b}, I_{HR}^{b},  N^{b} \right \}$ via ${D}_b$ as same as  $\left \{ I_{LR}^{a}, I_{HR}^{a},  N^{a} \right \}$
        \STATE Construct $\left \{ I_{LR}^{a+b}, I_{HR}^{a+b},  N^{a+b} \right \}$
        
        \end{algorithmic}
\end{algorithm}

\begin{figure*}
    \centering
    \includegraphics[width=0.93\textwidth]{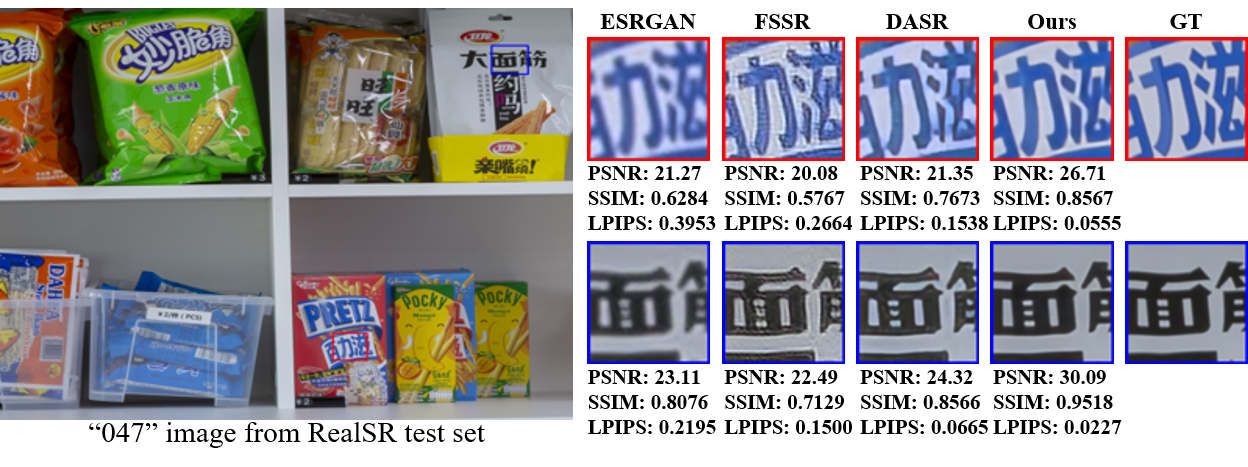} 
    \caption{Super-resolution results on the RealSR~\cite{realsr} dataset.}
    \label{fig:realsr}
\end{figure*}

\begin{figure*}
    \centering
    \includegraphics[width=0.93\textwidth]{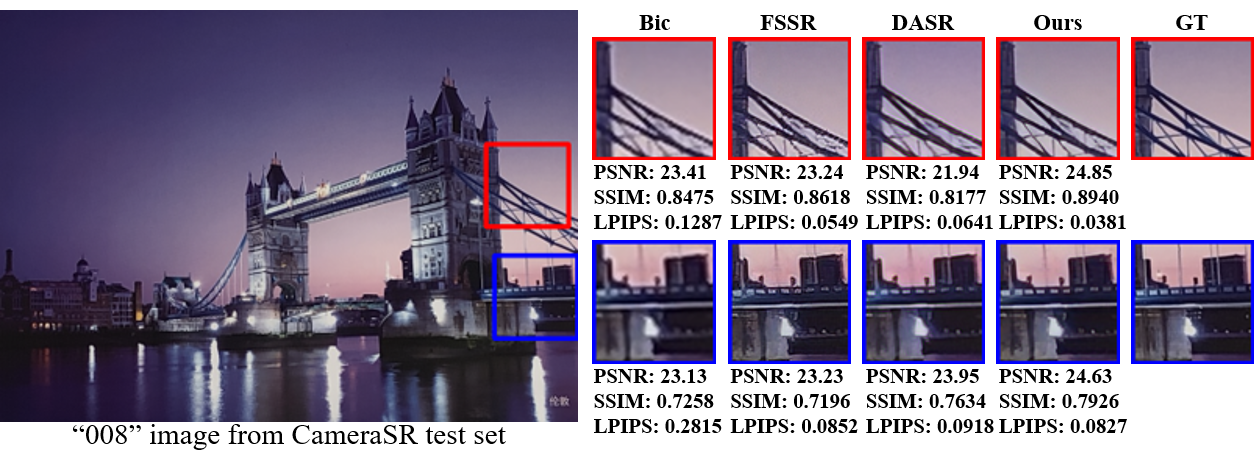} 
    \caption{Super-resolution results on the CameraSR~\cite{CameraSR} dataset.}
    \label{fig:camerasr}
\end{figure*}

\begin{figure*}
    \centering
    \includegraphics[width=0.93\textwidth]{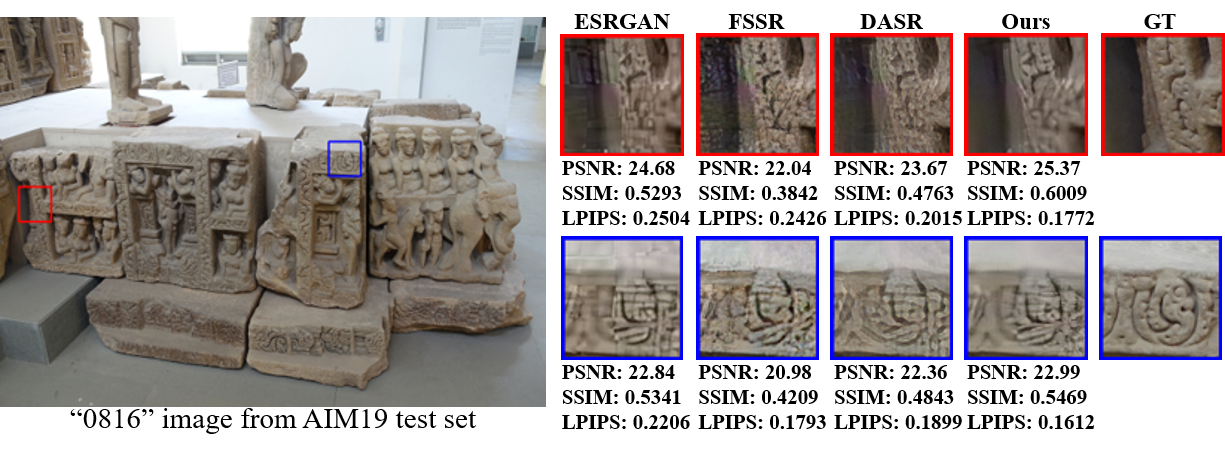}
    \caption{Super-resolution results on the AIM19~\cite{AIM19} challenge data.}
    \label{fig:aim19}
\end{figure*}

\section{Experiments}
\subsection{Dataset and Evaluation Protocols}
We use following four real-world SR datasets for comprehensive comparsions to validate our RWSR-EDL:
\begin{itemize}
  \item \emph{RealSR}~\cite{realsr} 
 consists of 595 LR-HR image pairs, which are collected under a real-world scenario with various optical resolution. To stay in step with other baseline methods~\cite{wei2020unsupervised,fritsche2019frequency}, we use 200 RealSR image pairs and 800 clean images from DIV2K for training. Then, we conduct specified 50 LR-HR pairs, which are collected by Canon camera, for testing. Due to the images from the RealSR dataset are captured under different optical settings, they inherently have different resolutions, we adopt $\times$4 scale to evaluate our RWSR-EDL.
 
  \item \emph{CameraSR}~\cite{CameraSR} is a real-world SR dataset, which consists of 200 LR-HR pairs collected by iPhoneX and Nikon camera, respectively. In our experiment, we used 80 real-world images, which are collected by iPhoneX (e.g., No.021-100) and 800 clean images from DIV2K for training. The rest 20 pairs of LR-HR pairs (No.001-020) are incorporated for testing.
  
  \item \emph{NTIRE2020 Challenge}~\cite{Lugmayr2020ntire} contains 3550 images, which downscaled with unknown noise to simulating inherent optical sensor noise. In our experiment, we use 3450 images, which consists of 2650 images from Flickr2K and 800 images from DIV2K, for training. The testing data contains 100 images from the DIV2K validation set with the same degradation operation as the training image. We adopt the $\times$4 scale to evaluate our RWSR-EDL.
  
  \item \emph{AIM2019 Challenge}~\cite{AIM19}. The training and testing images provided by AIM2019 are the same as NTIRE2020. Nevertheless, different and undisclosed downgrade metrics were used to generate corrupted input. We adopt the $\times$4 scale to evaluate our RWSR-EDL as well.
\end{itemize}

In evaluation protocols, we adopted Peak Signal-to-Noise Ratio (PSNR), Structural Similarity (SSIM), Learned Perceptual Image Patch Similarity (LPIPS)~\cite{lpips}, Naturalness Image Quality Evaluator (NIQE)~\cite{mittal2012making} and Mean Opinion Score (MOS) to verify RWSR-EDL. Meanwhile, PSNR and SSIM are the most commonly used evaluation metrics in image restoration, as they focus on pixel fidelity rather than human perception. In contrast, LPIPS and NIQE pays more attention to human visual perception, the lower values of LPIPS and NIQE score indicate beeter perception quality in the SR observations. We collect the MOS results by recruiting 20 volunteers for subjective assessment on NTIRE2020 challenge data. Specifically, 100 pairs of patches from the NTIRE2020 testset are randomly partitioned and the volunteers were shown a side-by-side comparison of each method's result and the referenced HR image. They were then asked to evaluate the quality of the SR image w.r.t. the reference image using the 6-level scale defined as: 0 - `Perfect’, 1 - `Almost Perfect’, 2 - `Slightly Worse’, 3 - `Worse’, 4 - `Much Worse’, 5 - `Terrible’.

\subsection{Implementation Details and Competing Methods}
In our experiments, we use flip and random rotation with angles of $90^{\circ}$, $180^{\circ}$ and $270^{\circ}$ for data augmentation. The images are cropped into $128 \times 128$ are input and the batchsize is 16. We use Adam~\cite{adam} as optimizer, where $\beta_1 =  0.9$ and $\beta_2 = 0.999$. The initial learning rate is $1 \times 10^{-4}$ and our RWSR-EDL is trained on an NVIDIA Tesla V100 server. Since the LR and HR images in the CameraSR dataset have the same resolution, we removed the corresponding downsampling and upsampling layers in CameraSR experiment. For the NGDC strategy, we adopt data of NTIRE2020 challenge~\cite{Lugmayr2020ntire} as an auxiliary dataset to present an efficient mix training in all experiments except the NTIRE2020 experiment itself.

Also, we incorporating ZSSR~\cite{zeroshot}, CinCGAN~\cite{CinCGAN}, ESRGAN~\cite{wang2018esrgan}, FSSR~\cite{fritsche2019frequency}, DASR~\cite{wei2020unsupervised}, Impressionism~\cite{Ji_2020_CVPR_Workshops}, FASRGAN~\cite{yan2021fine} and KernelGAN~\cite{kernelgan} on corresponding RealSR benchmark for comparison. For ZSSR~\cite{zeroshot}, DASR~\cite{wei2020unsupervised} and Impression~\cite{Ji_2020_CVPR_Workshops}, we adopt official models for evaluation, which are trained on corresponding dataset. To keep consistent with other baselines, we adopt pre-trained model of ESRGAN~\cite{wang2018esrgan} for testing. For FSSR~\cite{fritsche2019frequency}, we fine-tune its model on the corresponding datasets and obtain the results.

\def\tablename{Table}

\begin{table}[ht]
\centering
\begin{tabular}{l|ccc}
\hline
Methods     & \multicolumn{1}{l}{PSNR$\uparrow$}            & \multicolumn{1}{l}{SSIM$\uparrow$}   & \multicolumn{1}{l}{LPIPS$\downarrow$} \\ \hline \hline
ZSSR~\cite{zeroshot}        & 26.007 & 0.7482 & 0.386 \\
ESRGAN~\cite{wang2018esrgan} & 25.956          & 0.7468 & 0.415 \\
CinCGAN~\cite{CinCGAN}     & 25.094                              & 0.7459           & 0.405                     \\
FSSR~\cite{fritsche2019frequency}       & 25.992                              & 0.7388                     & 0.265                     \\ 
Impressionism~\cite{Ji_2020_CVPR_Workshops}        & 25.781        & 0.7508      & 0.258                  \\
FASRGAN~\cite{yan2021fine}        & 26.011        & 0.7504      & 0.307   \\
DASR~\cite{wei2020unsupervised}     & \underline{26.229}                              & \underline{0.7660}                     & \underline{0.251}           \\
 \hline \hline           
RWSR-EDL       & \textbf{27.803}          & \textbf{0.8112} & \textbf{0.247} \\ \hline
\end{tabular}
\caption{Quantitative results on RealSR dataset.}
\label{tab:realsr}
\end{table}

\begin{table}[ht]
\centering
\resizebox{\textwidth}{4.5mm}{
\begin{tabular}{l|cccccc}
\hline
Methods  &ZSSR &ESRGAN &CinCGAN &FSSR &DASR  & RWSR-EDL         \\ \hline \hline
NIQE & 4.971 & 6.327 &4.218 & 3.428 & 2.971 &2.419 \\ \hline
\end{tabular}}
\caption{Evaluation on RealSR~\cite{realsr} testset with NIQE index~\cite{mittal2012making}.}
\label{tab:niqe}
\end{table}

\begin{table}[ht]
\centering
\begin{tabular}{l|ccc}
\hline
Methods     & \multicolumn{1}{l}{PSNR$\uparrow$}            & \multicolumn{1}{l}{SSIM$\uparrow$}   & \multicolumn{1}{l}{LPIPS$\downarrow$} \\ \hline \hline
FSSR~\cite{fritsche2019frequency}        & 23.781                              & 0.7566                     & 0.180    \\
Impressionism~\cite{Ji_2020_CVPR_Workshops}        & 25.142              & \underline{0.8097}            & \underline{0.139}                    \\
DASR~\cite{wei2020unsupervised}       & \underline{25.235}                              & 0.8065                     & 0.141                   \\ \hline \hline
RWSR-EDL       & \textbf{26.284}          & \textbf{0.8226} & \textbf{0.133} \\ \hline
\end{tabular}
\caption{Quantitative results on CameraSR dataset.}
\label{tab:camerasr}
\end{table}

\begin{table}[ht]
\centering
\begin{tabular}{l|ccc}
\hline
Methods     & \multicolumn{1}{l}{PSNR$\uparrow$}            & \multicolumn{1}{l}{SSIM$\uparrow$}   & \multicolumn{1}{l}{LPIPS$\downarrow$} \\ \hline \hline
ZSSR~\cite{zeroshot}        & \underline{22.327}       & 0.6022        & 0.630 \\
ESRGAN~\cite{wang2018esrgan} & 21.382                & 0.5478        & 0.543 \\
CinCGAN~\cite{CinCGAN}    & 21.602                              & \underline{0.6129}            & 0.461                     \\
FSSR~\cite{fritsche2019frequency}        & 20.820                              & 0.5103                     & 0.390   \\
Impressionism~\cite{Ji_2020_CVPR_Workshops}        & 21.021                              & 0.5978                     & 0.376                     \\
DASR~\cite{wei2020unsupervised}       & 21.780                              & 0.5725                     & \underline{0.346}                   \\ \hline \hline
RWSR-EDL       & \textbf{22.335}                & \textbf{0.6187}        & \textbf{0.342} \\ \hline
\end{tabular}
\caption{Quantitative results of AIM2019 Challenge on Real-world image SR track. Note that FSSR is the champion method in AIM2019 Challenge.}
\label{tab:aim19}
\end{table}

\begin{table}[ht]
\centering
\begin{tabular}{l|cccc}
\hline                                   
Methods     & \multicolumn{1}{l}{PSNR$\uparrow$} & \multicolumn{1}{l}{SSIM$\uparrow$} & \multicolumn{1}{l}{LPIPS$\downarrow$} & \multicolumn{1}{l}{MOS$\downarrow$} \\ \hline \hline
EDSR~\cite{edsr}      & \underline{25.31}           & 0.6383                   & 0.5784      &2.875              \\
ESRGAN~\cite{wang2018esrgan} & 19.06                    & 0.2423                   & 0.7552      &3.250              \\
ZSSR~\cite{zeroshot}       & 25.13                    & 0.6268                   & 0.6160       &2.905              \\
KernelGAN~\cite{kernelgan}      & 18.46                    & 0.3826                   & 0.7307      & 3.155             \\
FASRGAN~\cite{yan2021fine}        & 21.86                              & 0.6214                     & 0.5499       &2.740    \\
Impressionism~\cite{Ji_2020_CVPR_Workshops}      & 24.82                    &\underline{0.6619}                   & \underline{0.2270}          &2.430           \\\hline \hline
RWSR-EDL        & \textbf{25.40}                   & \textbf{0.6819}          & \textbf{0.2222}    &2.225        \\ \hline
\end{tabular}
\caption{Quantitative results for NTIRE2020 Challenge on Real-world image SR track. Note that Impressionism is the winning approach in the NTIRE2020 Challenge. }
\label{tab:ntire20}
\end{table}

\begin{table}[ht]
\footnotesize
\begin{tabular}{l|cccc}
\hline
\multicolumn{1}{c|}{Method}  & ESRGAN & FSSR & DASR  & RWSR-EDL         \\ \hline \hline
\multicolumn{1}{c|}{Time(frame/s)} & 0.7971 & 0.7918 & \textbf{0.7465} & \underline{0.7632} \\ \hline
\multicolumn{1}{c|}{Parameter} & 16,697,987 & 16,697,987 & 16,697,987 & 16,729,694 \\ \hline
\end{tabular}
\caption{Efficiency analysis on 300$\times$200 image of RealSR~\cite{realsr} testset with $4 \times$ factor. As FSSR and DASR adopt ESRGAN as backbone, they have same network parameters.}
\label{tab:efficiency}
\end{table}

\begin{table}[ht]
\begin{tabular}{l|ccc}
\hline
\multicolumn{1}{c|}{Test Set} & \multicolumn{3}{c}{RealSR~\cite{realsr}}                                        \\ \hline
\multicolumn{1}{c|}{Metric}   & \multicolumn{1}{l}{PSNR$\uparrow$}   & \multicolumn{1}{l}{LPIPS$\downarrow$}   & \multicolumn{1}{l}{Parameter} \\ \hline \hline
Single Branch  & \underline{27.691} &  0.255& 16,708,556 \\
Dual-Learning w/o Mask    & 27.688 & \underline{0.254}& 16,719,125 \\
Dual-Learning w/ Mask     & \textbf{27.775} & \textbf{0.250} & 16,729,694 \\ \hline
\end{tabular}
\caption{Ablation study on different branches. With similar parameter number, `Dual-Learning w/ Mask' exhibits a significant improvement.}
\label{Tab:branches}
\end{table}

\begin{table}[ht]
\begin{tabular}{l|ccc}
\hline
\multicolumn{1}{c|}{Test Set} & \multicolumn{3}{c}{RealSR~\cite{realsr}}                                        \\ \hline
\multicolumn{1}{c|}{Metric}   & \multicolumn{1}{l}{PSNR$\uparrow$}   & \multicolumn{1}{l}{SSIM$\uparrow$}   & \multicolumn{1}{l}{LPIPS$\downarrow$} \\ \hline \hline
VGG-128  & 27.673 & 0.8058 & 0.254 \\
Patch-D    &\textbf{27.775} & \textbf{0.8095} & \textbf{0.250} \\ \hline
\end{tabular}
\caption{Ablation study on discriminator.}
\label{Tab:discriminator}
\end{table}

\begin{table}[ht]
\footnotesize
\begin{tabular}{l|ccc}
\hline
\multicolumn{1}{c|}{Test Set} & \multicolumn{3}{c}{RealSR~\cite{realsr}}                                        \\ \hline
\multicolumn{1}{c|}{Metric}   & \multicolumn{1}{l}{PSNR$\uparrow$}   & \multicolumn{1}{l}{SSIM$\uparrow$}   & \multicolumn{1}{l}{LPIPS$\downarrow$} \\ \hline \hline
Ours(RealSR) w/o Noise Sampling  & 27.715 & 0.8078 & 0.252 \\
Ours(RealSR) w/ Data Augmentation     & \underline{27.775} & \underline{0.8095} & \underline{0.250} \\
Ours(RealSR + NTIRE2020) w/o NGDC    & 27.557 & 0.8022 & 0.264 \\
Ours(RealSR + NTIRE2020) w/ NGDC     & \textbf{27.802} & \textbf{0.8110} & \textbf{0.247} \\ \hline
\end{tabular}
\caption{Ablation study on Noise-Guidance Data Collection (NGDC) strategy. With an auxiliary dataset, NGDC can consistently promote performance in all evaluation metrics without any training time increment.}
\label{Tab:NGDC}
\end{table}

\begin{table}[ht]
\begin{tabular}{l|ccc}
\hline
\multicolumn{1}{c|}{Test Set} & \multicolumn{3}{c}{RealSR~\cite{realsr}}                                        \\ \hline
\multicolumn{1}{c|}{Metric}   & \multicolumn{1}{l}{PSNR$\uparrow$}   & \multicolumn{1}{l}{SSIM$\uparrow$}   & \multicolumn{1}{l}{LPIPS$\downarrow$} \\ \hline \hline
$\widetilde{\mathcal{L}}_{all}$ in Equ.~\ref{eqa:oriloss} & 26.127 &0.7551 &0.272 \\
$\mathcal{L}_{total}=\mathcal{L}_{pix}$  & 27.690 & 0.8098 & 0.256 \\
$\mathcal{L}_{total}=\mathcal{L}_{pix} + \mathcal{L}_{per}$   &\underline{27.730} &\underline{0.8101} & \underline{0.251} \\
$\mathcal{L}_{total}=\mathcal{L}_{pix} + \mathcal{L}_{adv}$   &27.504 &0.8006 & 0.254 \\ $\mathcal{L}_{total}=\mathcal{L}_{pix} + \mathcal{L}_{per} + \mathcal{L}_{adv}$  &\textbf{27.802} &\textbf{0.8110} & \textbf{0.247} \\
\hline
\end{tabular}
\caption{Ablation study on loss functions.}
\label{Tab:loss}
\end{table}

\subsection{Quantitative and Qualitative Comparisons}

\textbf{RealSR.} As depicted in Tab.~\ref{tab:realsr}, we compare the state-of-the-art methods on RealSR dataset. Compared with DASR, our RWSR-EDL achieves clear improvement over three evaluation metrics. For instance, as DASR is the recently proposed method, RWSR-EDL obviously surpasses DASR with 1.57 dB and 0.045 on PSNR and SSIM, which justify the effectiveness of RWSR-EDL. Compared with the zero-shot learning method, RWSR-EDL still achieves 1.76dB improvement. Similarly, as shown in Tab.~\ref{tab:niqe}, the quantitative performance in terms of NIQE indicates considerable improvment using RWSR-EDL compared to other models. Also, we present the visual comparison in Fig.~\ref{fig:realsr}. It can be observed that other one-fold structure methods obtain blurry results on the RealSR dataset. By contrast, RWSR-EDL obtains clear structure and sharp details, which verify the effectiveness of the exclusionary dual-learning mechanism and NGDC strategy.

\textbf{CameraSR.} In Tab.~\ref{tab:camerasr}, we present the quantitative comparison on CameraSR dataset. Compared with DASR, our model achieve 1.05 dB improvement. Besides, RWSR-EDL still obtains 0.008 improvements over the LPIPS index, which justifies that our soft-mask mechanism makes a good balance between L1- and perceptual- minimization. As shown in Fig.~\ref{fig:camerasr}, RWSR-EDL also presents a high-quality restoration with more details. For example, compared with DASR, our method recovers more clear lines of the bridge on the upper row of Fig.~\ref{fig:camerasr}, which indeed show that the soft-mask strategy preserves sharp structure with adversarial optimization. 

\textbf{AIM2019 Challenge.} As depicted in Tab.~\ref{tab:aim19}, we compare RWSR-EDL with state-of-the-arts on data of the AIM2019 challenge. It notes that FSSR is the winning entry on the AIM2019 challenge. Compared with FSSR, our RWSR-EDL achieves 0.55 dB and 0.046 gain with PSNR and SSIM index. Also, RWSR-EDL consistently achieves first place among three evaluation metrics, which fully justifies that our exclusionary dual-learning mechanism helps RWSR-EDL realizes effective spatial attention among the multi-task paradigm. 

\begin{figure*}
    \centering
    \includegraphics[width=0.93\textwidth]{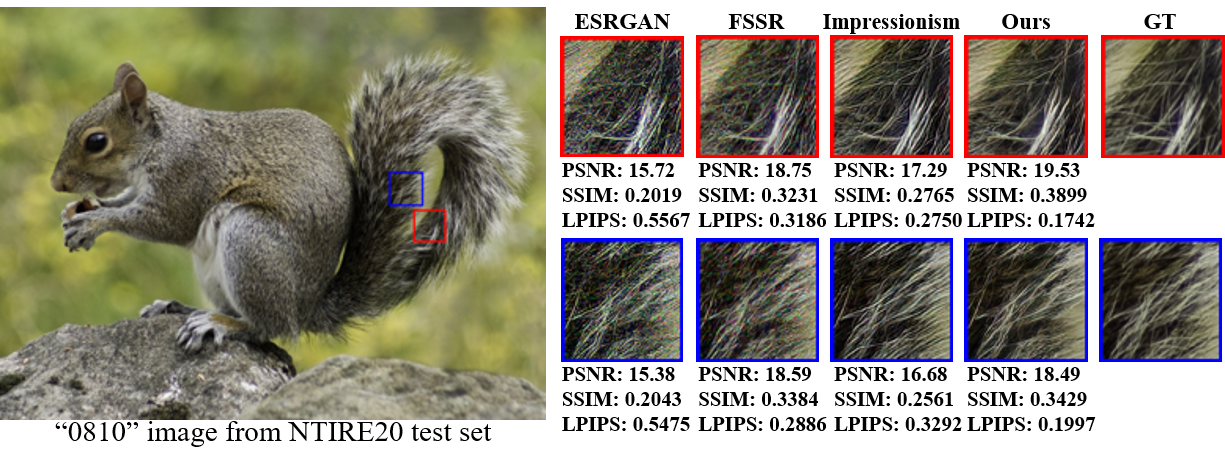} 
    \caption{Super-resolution results on the NTIRE2020~\cite{Lugmayr2020ntire} challenge data.}
    \label{fig:ntire20}
\end{figure*}

\textbf{NTIRE2020 Challenge.} In this experiment, we purely adopt data of NTIRE2020 and set the NGDC strategy down, to observe the effectiveness of our exclusionary dual-learning intuitively. As shown in Tab.~\ref{tab:ntire20}, RWSR-EDL achieves obvious improvement among three evaluation metrics. As Impressionism is the champion method in the NTIRE2020 challenge with high-quality enhancement, our model still performs superior results, which indeed verify the effectiveness of the proposed learning paradigm. Although EDSR enjoys much more parameter number and adopts L1 loss only, RWSR-EDL achieves 0.09 dB gain with PSNR index. Besides, EDSR exhibits a weak performance on perceptual-based evaluation protocol. Moreover, we confirmed the superior perceptual performance of RWSR-EDL by using MOS testing. The proposed model achieves the best result, with a 8.5\% better MOS than Impressionism. The qualitative results depicted in Fig.~\ref{fig:ntire20} verify that RWSR-EDL obtains significant visual quality improvement over Impressionism with realistic texture, clear structure, and fewer artifacts. 

\textbf{Efficiency Analysis.} Despite achieving superior results on quantitative comparisons, RWSR-EDL still presents a competitive running efficiency. The comparison in Tab.~\ref{tab:efficiency} reveals that RWSR-EDL achieves competing efficiency. Compared with ESRGAN, RWSR-EDL exhibits faster running efficiency and significant restoration quality improvement. Compared with DASR, our model performs obvious quantitative promotion with similar time consumption. 
 
\begin{figure*}
    \centering
    \includegraphics[width=0.93\textwidth]{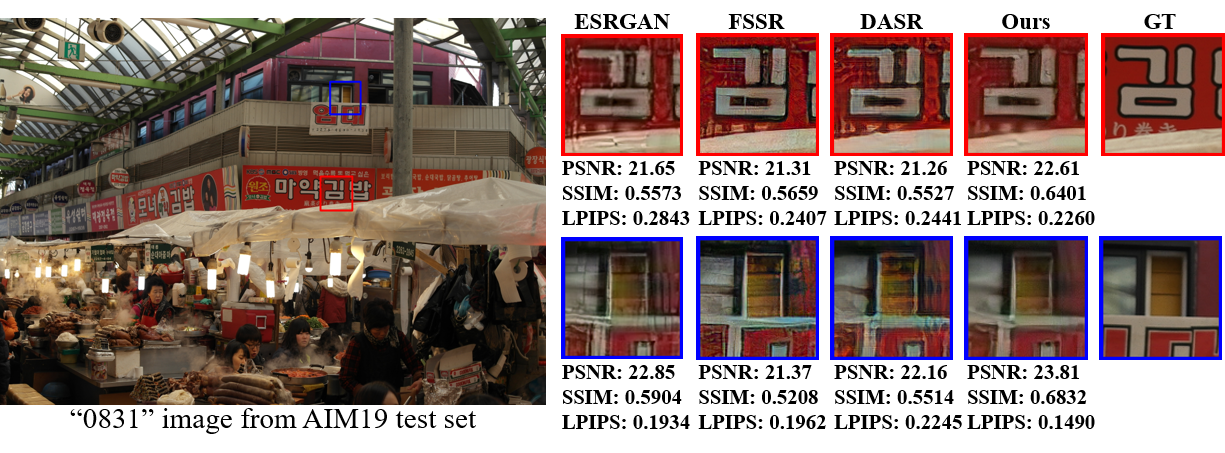}
    \caption{Super-resolution results on the AIM19~\cite{AIM19} challenge data.}
    \label{fig:aim19_2}
\end{figure*}

\begin{figure*}
    \centering
    \includegraphics[width=0.93\textwidth]{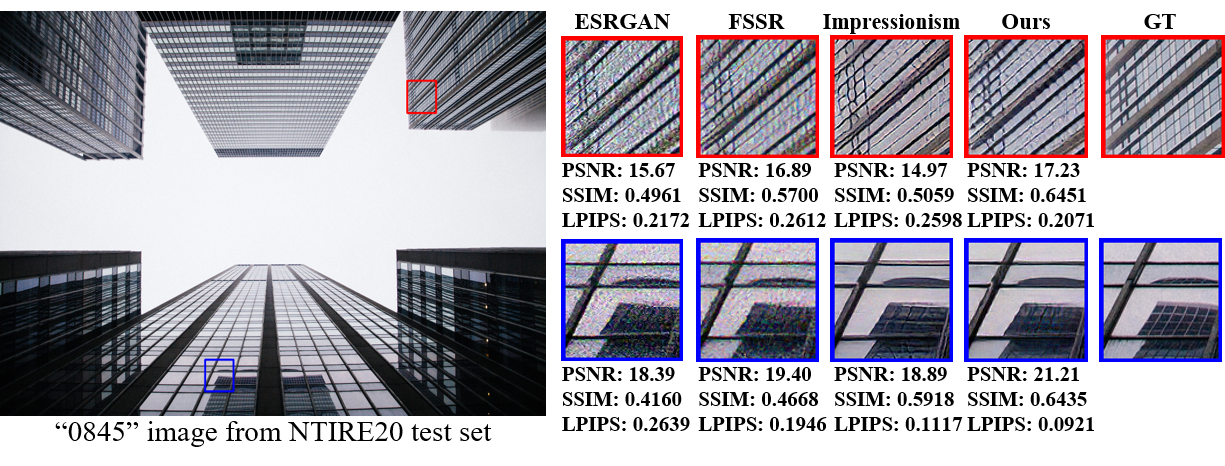} 
    \caption{Super-resolution results on the NTIRE2020~\cite{Lugmayr2020ntire} challenge data.}
    \label{fig:ntire20_2}
\end{figure*}
\subsection{Ablation Study}
In this section, we make an ablation study on the proposed components, which are exclusionary dual-learning, patch-discriminator, and NGDC strategy.

\textbf{Exclusionary Dual-Learning.} We make a study on the exclusionary dual-learning mechanism to justify the effectiveness of the proposed exclusionary dual-learning strategy. As depicted in Tab.~\ref{Tab:branches}, `Single Branch' shows a similar performance with `Dual-Learning w/o Mask', which justifies that a simply dual-way structure is useless to RealSR yet bring significant parameter increment. However, a combination of exclusionary soft-mask and dual-learning structure exhibits an obvious improvement over PSNR and LPIPS metrics both. Without obvious parameter increment, `Dual-Learning w/ Mask' surpasses `Dual-Learning w/o Mask' with 0.1 dB improvement and 0.005 LPIPS gain, which clearly justify that the proposed exclusionary soft-mask mechanism is adequate to the dual-way learning.

\textbf{NGDC Strategy.} To fully justify the effectiveness of NGDC strategy, we plot the training curves with PSNR and LPIPS index in Fig.~\ref{fig:curves}. Specifically, we conduct the evaluation on the testset of RealSR during the training phrase to plot the curves. As depicted in Fig.~\ref{fig:curves}, `Ours(RealSR + NTIRE2020) w/ NGDC' is convergenced while the `Ours(RealSR + NTIRE2020) w/o NGDC' still stays at a climbing state, which fully justifies that the proposed NGDC strategy can extract useful noise from the auxiliary dataset and utilize noise stacking to bring superior performance. In Tab.~\ref{Tab:NGDC}, we also present a quantitative study on NGDC strategy. As noise sampling strategy give a competitive performance within a single dataset, they fail to demonstrate similar promotion when incorporating auxiliary large-scale dataset as the training data is too miscellaneous. For example, `Ours(RealSR + NTIRE2020) w/o NGDC' shows a lower score on all evaluation metrics than `Our(RealSR) w/ Data Augmentation'. To fair comparison, `Ours(RealSR + NTIRE2020) w/o NGDC' and `Ours(RealSR + NTIRE2020) w/ NGDC' both adopt data augmentation strategy as same as `Ours(RealSR) w/ Data Augmentation'. To our surprise, our NGDC strategy can significant improve the results when incorporating auxiliary large-scale dataset and noise sampling strategy, which well verify that auxiliary dataset exists a certain portion of negative samples and noise-guidance data collection is necessary.

\textbf{Patch Discriminator.} We make an experiment on the discriminator of the adversarial training. As shown in Tab.~\ref{Tab:discriminator}, `Patch-D' achieves 27.75 dB and the `VGG-128' obtains 26.67 dB, which justify the effectiveness of patch discriminator in real-world image enhancement. Moreover, patch discriminator also brings a significant improvement on SSIM and LPIPS. To this end, we replace VGG-128 with patch discriminator as default discriminator in the adversarial optimization.

\textbf{Loss Function.} We compare different settings of loss on the RealSR dataset. As depicted in Tab.~\ref{Tab:loss}, our model achieves promising results on PSNR by incorporating $\mathcal{L}_{per}$ and $\mathcal{L}_{pix}$. Compared with the typical weighted sum loss $\widetilde{\mathcal{L}}_{all}$ in Equ.~\ref{eqa:oriloss}, our method achieves 1.68 dB and 0.025 gain with PSNR and LPIPS index. However, the proposed model shows a weaker PSNR score when incorporating the adversarial loss function. To our surprise, our full model achieves further improvement as the adversarial loss is incorporated, which justify that $\mathcal{L}_{per}$ is important in RealSR issue. Also, our full model achieves the best score among Tab.~\ref{Tab:loss} verify the effectiveness of the proposed exclusionary dual-learning in L1- and perceptual- based cooperative optimization.

\section{Conclusion and Future Work}
In this work, we revival exclusionary dual-learning to facilitate deep representation that exhibits more diversity under L1- and perceptual- minimization. With the novel exclusionary dual-learning mechanism, our RWSR-EDL demonstrates a charming result on real-world image super-resolution. Moreover, we present a noise-guidance data collection strategy, which yields an efficient training paradigm in multiple datasets learning. Experimental results show that RWSR-EDL surpasses state-of-the-art real-world image super-resolution methods with a clear margin and closed parameter number on perceptual- and euclidean- based evaluation protocols. 

We aim at extending our work by following directions. First, we would like to extend the proposed NGDC strategy to a real-world image denoising task, aim at collecting more noise samples with a given image and demonstrate efficient denoising training. Second, we are considering involving explainable deep learning into our framework to further improve the exclusionary dual-learning mechanism.


%

\bibliographystyle{IEEEtran}
\bibliography{egbib}



\end{document}